\documentclass{article}

\usepackage[preprint,nonatbib]{neurips_2023}
\usepackage{graphicx}

\usepackage[utf8]{inputenc} 
\usepackage[T1]{fontenc}    
\usepackage[colorlinks,
    colorlinks=true, 
    linkcolor=blue, 
    filecolor=blue, 
    citecolor=blue,       
    urlcolor=blue]{hyperref}       
\usepackage{url}            
\usepackage{booktabs}       
\usepackage{amsfonts}       
\usepackage{amsmath}
\usepackage{amsthm}
\usepackage{amssymb}
\usepackage{nicefrac}       
\usepackage{microtype}      
\usepackage{xcolor}         
\usepackage[font=small,labelfont=bf]{caption}
\usepackage{enumitem}
\usepackage{wrapfig}
\usepackage{listings}
\usepackage{caption}

\usepackage[normalem]{ulem}
\usepackage{xspace}
\usepackage{float}
\usepackage{tabularx}
\usepackage{booktabs}
\usepackage{multirow}
\usepackage[normalem]{ulem}
\usepackage[export]{adjustbox}
\useunder{\uline}{\ul}{}

\def\llama{Llama\xspace}
\def\mistralSB{Mistral~7B\xspace}
\def\mixtral{Mixtral\xspace}
\def\mixtralEXSB{Mixtral~8x7B\xspace}
\def\mixtralchat{Mixtral~--~Instruct\xspace}
\def\mixtralEXSBchat{Mixtral~8x7B~--~Instruct\xspace}

\title{Mixtral of Experts}

\author{%
Albert Q. Jiang, Alexandre Sablayrolles, Antoine Roux, Arthur Mensch,\\
\textbf{Blanche Savary, Chris Bamford, Devendra Singh Chaplot, Diego de las Casas,} \\
\textbf{ Emma Bou Hanna, Florian Bressand, Gianna Lengyel, Guillaume Bour,} \\
\textbf{ Guillaume Lample, Lélio Renard Lavaud, Lucile Saulnier, Marie-Anne Lachaux,} \\
\textbf{ Pierre Stock, Sandeep Subramanian, Sophia Yang, Szymon Antoniak, Teven Le Scao,} \\
\textbf{ Théophile Gervet, Thibaut Lavril, Thomas Wang, Timothée Lacroix, William El Sayed } \\
}
\begin{document}

\maketitle
\begin{center}
\vspace{-30pt}
\centering
\includegraphics[width=0.8\linewidth,keepaspectratio]{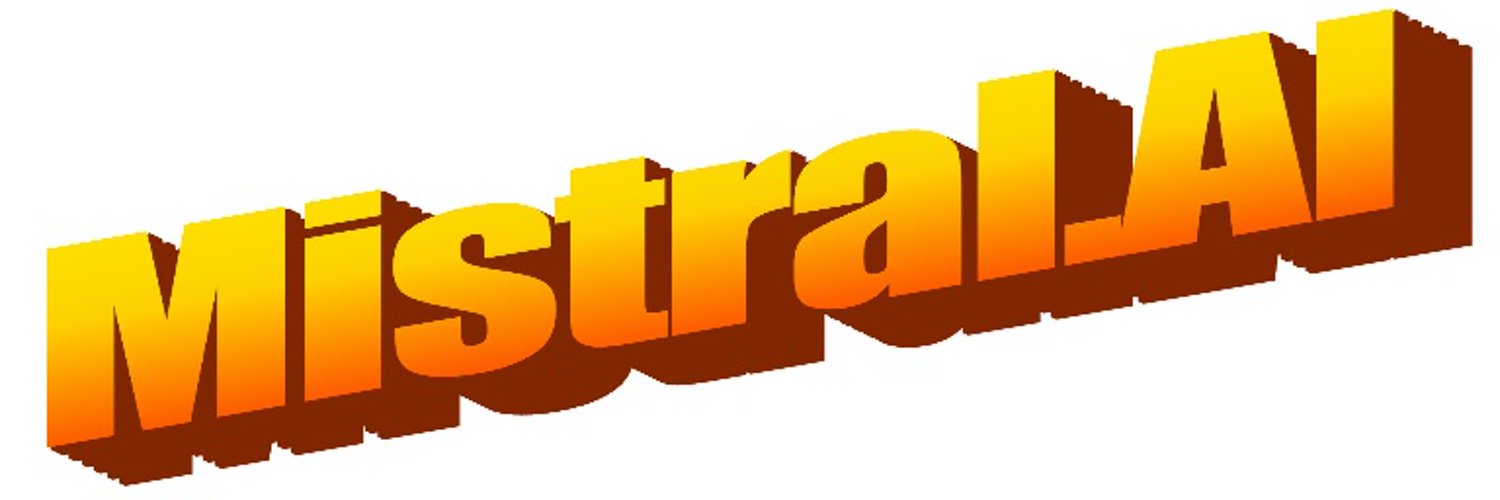}
\end{center}

\begin{abstract}

We introduce \mixtralEXSB, a Sparse Mixture of Experts (SMoE) language model.
\mixtral has the same architecture as \mistralSB, with the difference that each layer is composed of 8 feedforward blocks (i.e. experts).
For every token, at each layer, a router network selects two experts to process the current state and combine their outputs.
Even though each token only sees two experts, the selected experts can be different at each timestep.
As a result, each token has access to 47B parameters, but only uses 13B active parameters during inference. 
\mixtral was trained with a context size of 32k tokens and it outperforms or matches \llama~2~70B and GPT-3.5 across all evaluated benchmarks.
In particular, \mixtral vastly outperforms \llama~2~70B on mathematics, code generation, and multilingual benchmarks.
We also provide a model fine-tuned to follow instructions, \mixtralEXSBchat, that surpasses GPT-3.5 Turbo, Claude-2.1, Gemini Pro, and \llama~2~70B~--~chat model on human benchmarks. 
Both the base and instruct models are released under the Apache 2.0 license. \\

\vspace{-5pt}
\textbf{Code:} \url{https://github.com/mistralai/mistral-src} \\
\textbf{Webpage:} \url{https://mistral.ai/news/mixtral-of-experts/}

\end{abstract}

\section{Introduction}
\vspace{-5pt}
In this paper, we present \mixtralEXSB, a sparse mixture of experts model (SMoE) with open weights, licensed under Apache 2.0. \mixtral outperforms \llama 2 70B and GPT-3.5 on most benchmarks. As it only uses a subset of its parameters for every token, \mixtral allows faster inference speed at low batch-sizes, and higher throughput at large batch-sizes.

\mixtral is a sparse mixture-of-experts network. It is a decoder-only model where the feedforward block picks from a set of 8 distinct groups of parameters. At every layer, for every token, a router network chooses two of these groups (the “experts”) to process the token and combine their output additively. This technique increases the number of parameters of a model while controlling cost and latency, as the model only uses a fraction of the total set of parameters per token.

\mixtral is pretrained with multilingual data using a context size of 32k tokens. It either matches or exceeds the performance of \llama 2 70B and GPT-3.5, over several benchmarks. In particular, \mixtral demonstrates superior capabilities in mathematics, code generation, and tasks that require multilingual understanding, significantly outperforming \llama 2 70B in these domains. Experiments show that \mixtral is able to successfully retrieve information from its context window of 32k tokens, regardless of the sequence length and the location of the information in the sequence.

We also present \mixtralEXSBchat, a chat model fine-tuned to follow instructions using supervised fine-tuning and Direct Preference Optimization~\cite{rafailov2023direct}. Its performance notably surpasses that of GPT-3.5 Turbo, Claude-2.1, Gemini Pro, and \llama 2 70B – chat model on human evaluation benchmarks.
\mixtralchat also demonstrates reduced biases, and a more balanced sentiment profile in benchmarks such as BBQ, and BOLD.

We release both \mixtralEXSB and \mixtralEXSBchat under the Apache 2.0 license\footnote{\url{https://mistral.ai/news/mixtral-of-experts/}}, free for academic and commercial usage, ensuring broad accessibility and potential for diverse applications. 
To enable the community to run \mixtral with a fully open-source stack, we submitted changes to the vLLM project, which integrates Megablocks CUDA kernels for efficient inference.
Skypilot also allows the deployment of vLLM endpoints on any instance in the cloud. 

\section{Architectural details}
\vspace{-5pt}
\begin{wrapfigure}{r}{0.275\textwidth}
\center
\small
\vspace{-15pt}
\begin{tabular}{lr}
\toprule
\textbf{Parameter}  & \textbf{Value} \\ \midrule
\texttt{dim}             & $4096$           \\
\texttt{n\_layers}       & $32$             \\
\texttt{head\_dim}       & $128$            \\
\texttt{hidden\_dim}     & $14336$          \\
\texttt{n\_heads}        & $32$             \\
\texttt{n\_kv\_heads}    & $8$              \\
\texttt{context\_len} & $32768$           \\
\texttt{vocab\_size}     & $32000$          \\ 
\texttt{num\_experts}  & $8$ \\
\texttt{top\_k\_experts}  & $2$ \\
\bottomrule
\end{tabular}
\vspace{-5pt}
\captionof{table}{\small \textbf{Model architecture.}}
\label{tab:param}
\vspace{-25pt}
\end{wrapfigure}

\mixtral is based on a transformer architecture~\cite{vaswani2017attention} and uses the same modifications as described in \cite{jiang2023mistral}, with the notable exceptions that \mixtral supports a fully dense context length of 32k tokens, and the feedforward blocks are replaced by Mixture-of-Expert layers (Section~\ref{sec:smoe}).
The model architecture parameters are summarized in Table~\ref{tab:param}.

\subsection{Sparse Mixture of Experts}
\label{sec:smoe}
\vspace{-5pt}

\begin{figure*}
\centering
\vspace{-12pt}
\includegraphics[width=0.6\linewidth,keepaspectratio]{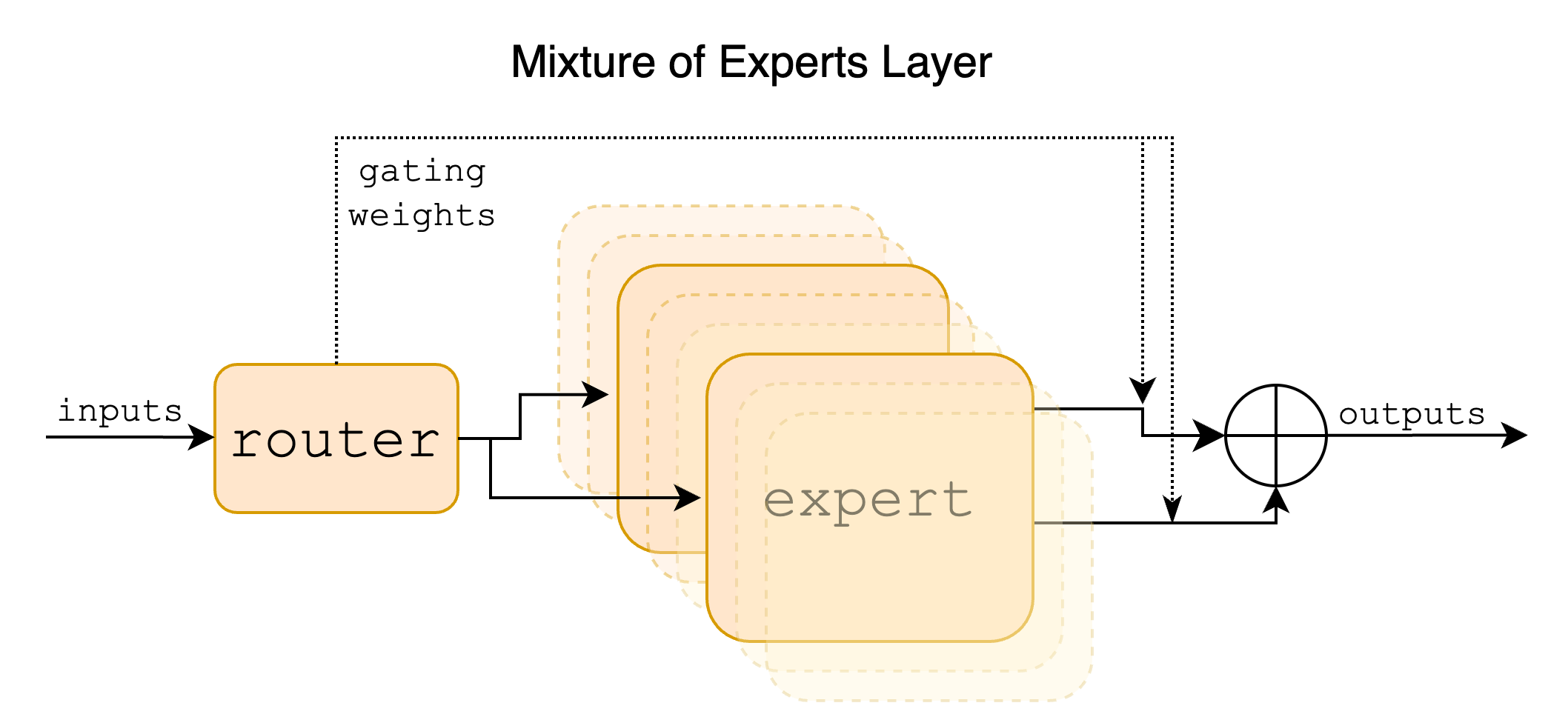}
\caption{\small \textbf{Mixture of Experts Layer.} Each input vector is assigned to 2 of the 8 experts by a router. The layer's output is the weighted sum of the outputs of the two selected experts.
In \mixtral, an expert is a standard feedforward block as in a vanilla transformer architecture.
}
\label{fig:smoe}
\vspace{-15pt}
\end{figure*}

We present a brief overview of the Mixture of Experts layer (Figure~\ref{fig:smoe}). 
For a more in-depth overview, see \cite{fedus2022review}.
The output of the MoE module for a given input \( x \) is determined by the weighted sum of the outputs of the expert networks, where the weights are given by the gating network's output. i.e. given $n$ expert networks \(\{E_0, E_i, ..., E_{n-1}\}\), the output of the expert layer is given by:
\vspace{-5pt}
\[ \sum_{i=0}^{n-1} G(x)_i \cdot E_i(x). \]

Here, \( G(x)_i \) denotes the \(n\)-dimensional output of the gating network for the \(i\)-th expert, and \( E_i(x) \) is the output of the \(i\)-th expert network. If the gating vector is sparse, we can avoid computing the outputs of experts whose gates are zero. There are multiple alternative ways of implementing $G(x)$~\cite{clark2022unified,hazimeh2021dselect,zhou2022mixture}, but a simple and performant one is implemented by taking the softmax over the Top-K logits of a linear layer~\cite{shazeer2017outrageously}. We use
\[ G(x) := \text{Softmax}(\text{TopK}(x \cdot W_g)), \]

where $(\text{TopK}(\ell))_i := \ell_i$ if $\ell_i$ is among the top-K coordinates of logits $\ell \in \mathbb{R}^n$ and $(\text{TopK}(\ell))_i := -\infty$ otherwise. 
The value of K -- the number of experts used per token -- is a hyper-parameter that modulates the amount of compute used to process each token.
If one increases $n$ while keeping $K$ fixed, one can increase the model's parameter count while keeping its computational cost effectively constant.
This motivates a distinction between the model's total parameter count (commonly referenced as the \textbf{sparse} parameter count), which grows with $n$, and the number of parameters used for processing an individual token (called the \textbf{active} parameter count), which grows with $K$ up to $n$. 

MoE layers can be run efficiently on single GPUs with high performance specialized kernels. For example, Megablocks~\cite{gale2022megablocks} casts the feed-forward network (FFN) operations of the MoE layer as large sparse matrix multiplications, significantly enhancing the execution speed and naturally handling cases where different experts get a variable number of tokens assigned to them.
Moreover, the MoE layer can be distributed to multiple GPUs through standard Model Parallelism techniques, and through a particular kind of partitioning strategy called Expert Parallelism (EP)~\cite{shazeer2017outrageously}.
During the MoE layer's execution, tokens meant to be processed by a specific expert are routed to the corresponding GPU for processing, and the expert's output is returned to the original token location.
Note that EP introduces challenges in load balancing, as it is essential to distribute the workload evenly across the GPUs to prevent overloading individual GPUs or hitting computational bottlenecks.

\looseness=-1 In a Transformer model, the MoE layer is applied independently per token and replaces the feed-forward (FFN) sub-block of the transformer block. For \mixtral we use the same SwiGLU architecture as the expert function $E_i(x)$ and set $K=2$. This means each token is routed to two SwiGLU sub-blocks with different sets of weights. Taking this all together, the output $y$ for an input token \( x \) is computed as:
\[ y = \sum_{i=0}^{n-1} \text{Softmax}(\text{Top2}(x \cdot W_g))_i \cdot \text{SwiGLU}_i(x). \]

This formulation is similar to the GShard architecture \cite{lepikhin2020gshard}, with the exceptions that we replace all FFN sub-blocks by MoE layers while GShard replaces every other block, and that GShard uses a more elaborate gating strategy for the second expert assigned to each token.

\section{Results}
\vspace{-5pt}
We compare \mixtral to \llama, and re-run all benchmarks with our own evaluation pipeline for fair comparison.
We measure performance on a wide variety of tasks categorized as follow:

\begin{figure*}[b]
\centering
\vspace{-10pt}
\includegraphics[width=\linewidth,height=\textheight,keepaspectratio]{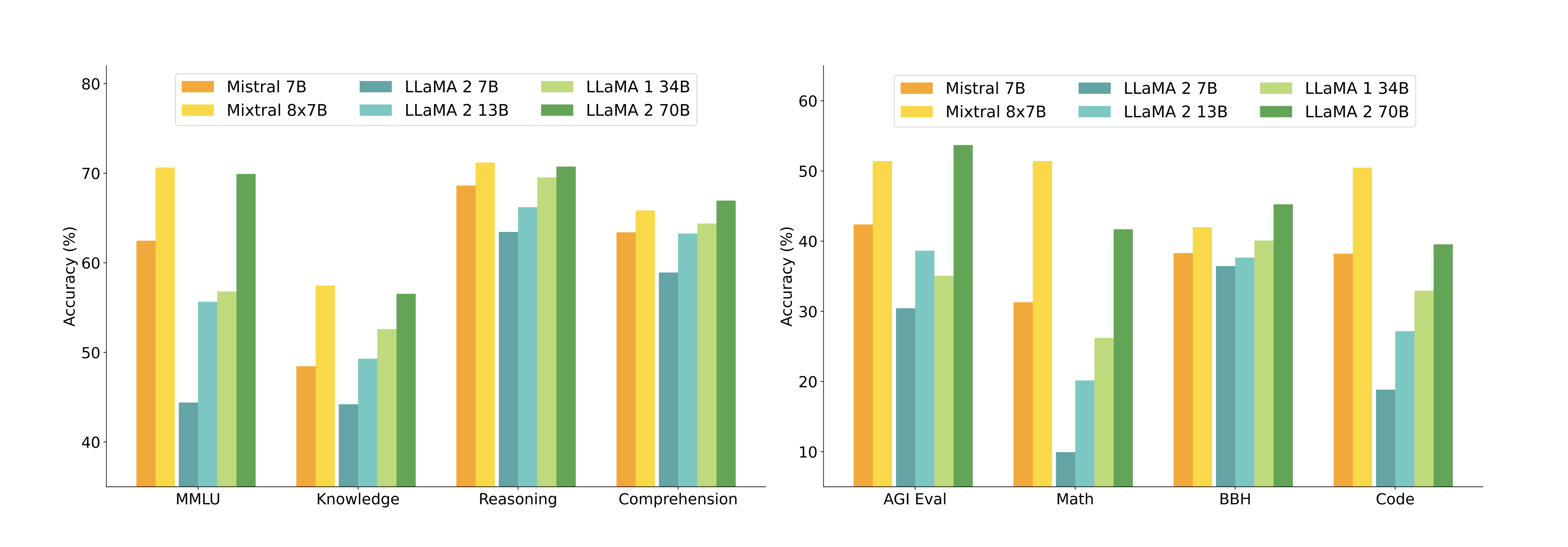}
\vspace{-20pt}
\caption{
\small
\textbf{Performance of \mixtral and different \llama models on a wide range of benchmarks}. All models were re-evaluated on all metrics with our evaluation pipeline for accurate comparison. \mixtral outperforms or matches \llama 2 70B on all benchmarks. In particular, it is vastly superior in mathematics and code generation.
}
\label{fig:bars}
\end{figure*}

\setlength{\tabcolsep}{0pt}
\begin{table}[t]
\scriptsize
\begin{tabular}{>{\arraybackslash}p{1.65cm}>{\centering\arraybackslash}p{0.9cm}*{12}{>{\centering\arraybackslash}p{0.95cm}}}
\toprule
\textbf{Model}        & \tiny{\textbf{\begin{tabular}[c]{@{}c@{}}Active \\ Params\end{tabular}}} & \textbf{MMLU}   & \textbf{HellaS} & \textbf{WinoG}  & \textbf{PIQA}   & \textbf{Arc-e}  & \textbf{Arc-c}  & \textbf{NQ}     & \textbf{TriQA} & \textbf{HumanE} & \textbf{MBPP}   & \textbf{Math}    & \textbf{GSM8K}  \\ \midrule
\textbf{LLaMA 2 7B}   & 7B                                                                & 44.4\%          & 77.1\%             & 69.5\%          & 77.9\%          & 68.7\%          & 43.2\%          & 17.5\%          & 56.6\%            & 11.6\%             & 26.1\%          & 3.9\%           & 16.0\%          \\[5pt]
\textbf{LLaMA 2 13B}  & 13B                                                               & 55.6\%          & 80.7\%             & 72.9\%          & 80.8\%          & 75.2\%          & 48.8\%          & 16.7\%          & 64.0\%            & 18.9\%             & 35.4\%          & 6.0\%           & 34.3\%          \\[5pt]
\textbf{LLaMA 1 33B}  & 33B                                                               & 56.8\%          & 83.7\%             & 76.2\%          & 82.2\%          & 79.6\%          & 54.4\%          & 24.1\%          & 68.5\%            & 25.0\%             & 40.9\%          & 8.4\%           & 44.1\%          \\[5pt]
\textbf{LLaMA 2 70B}  & 70B                                                               & 69.9\%          & \textbf{85.4\%}    & \textbf{80.4\%} & 82.6\%          & 79.9\%          & 56.5\%          & 25.4\%          & \textbf{73.0\%}   & 29.3\%             & 49.8\%          & 13.8\%          & 69.6\%          \\\midrule
\textbf{Mistral 7B}   & 7B                                                                & 62.5\%          & 81.0\%             & 74.2\%          & 82.2\%          & 80.5\%          & 54.9\%          & 23.2\%          & 62.5\%            & 26.2\%             & 50.2\%          & 12.7\%          & 50.0\%          \\[5pt]
\textbf{\mixtralEXSB} & 13B                                                               & \textbf{70.6\%} & 84.4\%             & 77.2\%          & \textbf{83.6\%} & \textbf{83.1\%} & \textbf{59.7\%} & \textbf{30.6\%} & 71.5\%            & \textbf{40.2\%}    & \textbf{60.7\%} & \textbf{28.4\%} & \textbf{74.4\%} \\ \bottomrule
\end{tabular}
\vspace{2pt}
\caption{
\small
\textbf{Comparison of \mixtral with \llama.} \mixtral outperforms or matches \llama 2 70B performance on almost all popular benchmarks while using 5x fewer active parameters during inference.
}
\label{tab:results}
\end{table}

\begin{itemize}[leftmargin=10pt]
\item \textbf{Commonsense Reasoning (0-shot):} Hellaswag~\cite{zellers2019hellaswag}, Winogrande~\cite{sakaguchi2021winogrande}, PIQA~\cite{bisk2020piqa}, SIQA~\cite{sap2019socialiqa}, OpenbookQA~\cite{mihaylov2018can}, ARC-Easy, ARC-Challenge~\cite{clark2018think}, CommonsenseQA~\cite{talmor2018commonsenseqa}
\item \textbf{World Knowledge (5-shot):} NaturalQuestions~\cite{kwiatkowski2019natural}, TriviaQA~\cite{joshi2017triviaqa}
\item \textbf{Reading Comprehension (0-shot):} BoolQ~\cite{clark2019boolq}, QuAC~\cite{choi2018quac}
\item \textbf{Math:} GSM8K~\cite{cobbe2021training} (8-shot) with maj@8 and MATH~\cite{hendrycks2021measuring} (4-shot) with maj@4
\item \textbf{Code:} Humaneval~\cite{chen2021evaluating} (0-shot) and MBPP~\cite{austin2021program} (3-shot)
\item \textbf{Popular aggregated results:} MMLU~\cite{hendrycks2020measuring} (5-shot), BBH~\cite{suzgun2022challenging} (3-shot), and AGI Eval~\cite{zhong2023agieval} (3-5-shot, English multiple-choice questions only)
\end{itemize}

Detailed results for \mixtral, \mistralSB and \llama 2 7B/13B/70B and \llama 1 34B\footnote{Since \llama 2 34B was not open-sourced, we report results for \llama 1 34B.} are reported in Table~\ref{tab:results}.
Figure~\ref{fig:bars} compares the performance of \mixtral with the \llama models in different categories.
\mixtral surpasses \llama 2 70B across most metrics.
In particular, \mixtral displays a superior performance in code and mathematics benchmarks.

\begin{figure*}
\centering
\includegraphics[width=0.85\linewidth,keepaspectratio]{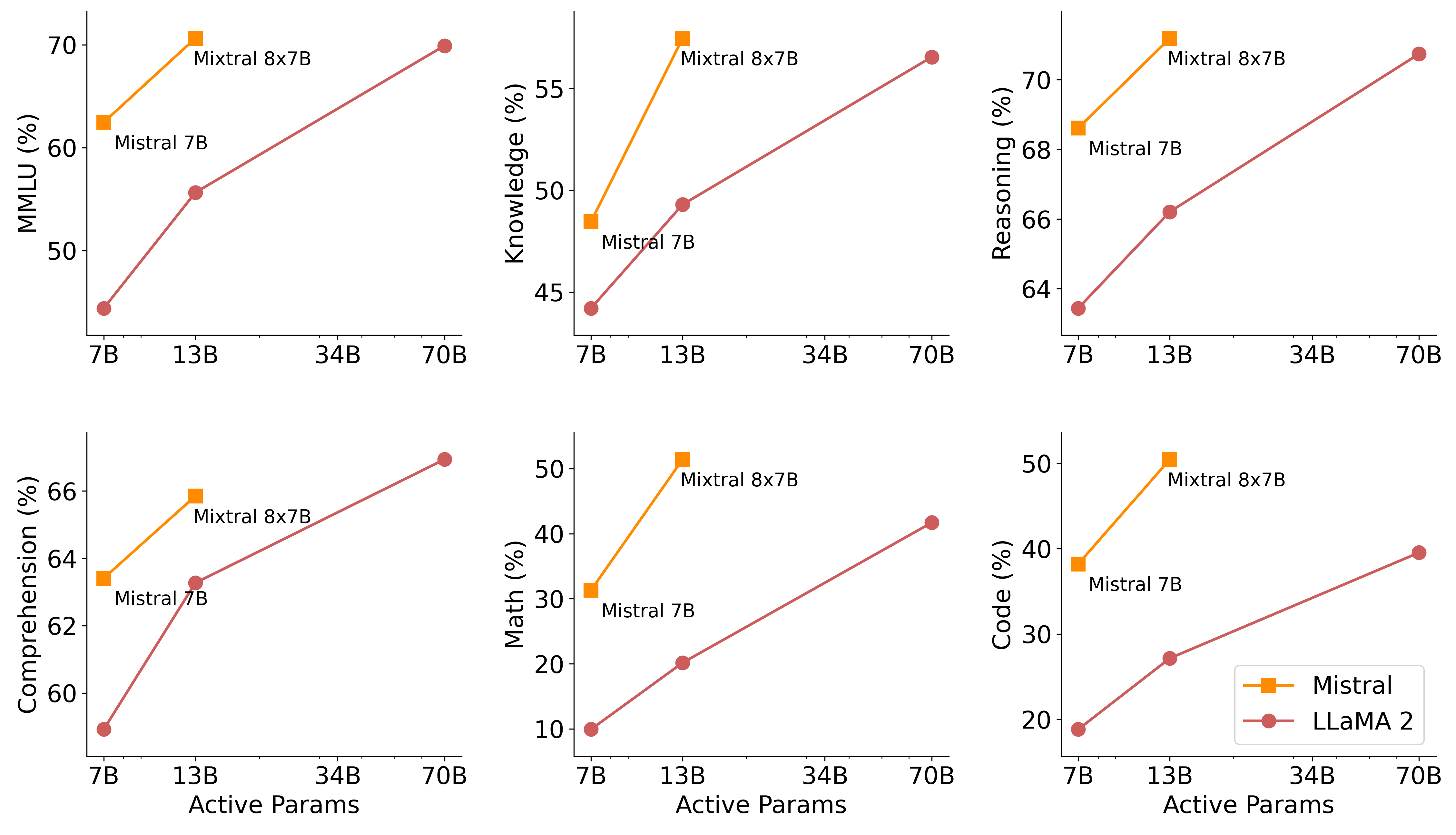}
\vspace{-5pt}
\caption{
\looseness=-1 \small \textbf{Results on MMLU, commonsense reasoning, world knowledge and reading comprehension, math and code for Mistral (7B/8x7B) vs \llama 2 (7B/13B/70B)}.
\mixtral largely outperforms \llama 2 70B on all benchmarks, except on reading comprehension benchmarks while using 5x lower active parameters.
It is also vastly superior to \llama 2 70B on code and math.
}
\label{fig:size}
\vspace{-10pt}
\end{figure*}

\looseness=-1 \textbf{Size and Efficiency.} We compare our performance to the \llama 2 family, aiming to understand \mixtral models' efficiency in the cost-performance spectrum (see Figure~\ref{fig:size}). As a sparse Mixture-of-Experts model, \mixtral only uses 13B active parameters for each token. With 5x lower active parameters, \mixtral is able to outperform \llama 2 70B across most categories.

Note that this analysis focuses on the active parameter count (see Section~\ref{sec:smoe}), which is directly proportional to the inference compute cost, but does not consider the memory costs and hardware utilization. 
The memory costs for serving \mixtral are proportional to its \emph{sparse} parameter count, 47B, which is still smaller than \llama 2 70B. As for device utilization, we note that the SMoEs layer introduces additional overhead due to the routing mechanism and due to the increased memory loads when running more than one expert per device. They are more suitable for batched workloads where one can reach a good degree of arithmetic intensity.

\textbf{Comparison with \llama 2 70B and GPT-3.5.} In Table~\ref{tab:vs_gpt35}, we report the performance of \mixtralEXSB compared to \llama 2 70B and GPT-3.5. We observe that \mixtral performs similarly or above the two other models.
On MMLU, \mixtral obtains a better performance, despite its significantly smaller capacity (47B tokens compared to 70B).
For MT Bench, we report the performance of the latest GPT-3.5-Turbo model available, \texttt{gpt-3.5-turbo-1106}.

\textbf{Evaluation Differences.} On some benchmarks, there are some differences between our evaluation protocol and the one reported in the \llama 2 paper: 1) on MBPP, we use the hand-verified subset 2) on TriviaQA, we do not provide Wikipedia contexts.

\setlength{\tabcolsep}{9pt}
\begin{table}
\small
\centering
\vspace{-20pt}
\begin{tabular}{@{}cccc@{}}
\toprule
\textbf{}                                                                & \textbf{LLaMA 2 70B} & \textbf{GPT-3.5} & \textbf{\mixtralEXSB} \\ \midrule
\begin{tabular}[c]{@{}c@{}}\textbf{MMLU}\\[-2pt] \scriptsize{ (MCQ in 57 subjects)}\end{tabular}      & 69.9\%               & 70.0\%             & \textbf{70.6\%}       \\[7pt]
\begin{tabular}[c]{@{}c@{}}\textbf{HellaSwag}\\[-2pt] \scriptsize{ (10-shot)}\end{tabular}            & \textbf{87.1}\%               & 85.5\%             & 86.7\%                \\[7pt]
\begin{tabular}[c]{@{}c@{}}\textbf{ARC Challenge}\\[-2pt] \scriptsize{ (25-shot)}\end{tabular}        & 85.1\%               & 85.2\%             & \textbf{85.8\%}       \\[7pt]
\begin{tabular}[c]{@{}c@{}}\textbf{WinoGrande}\\[-2pt] \scriptsize{ (5-shot)}\end{tabular}            & \textbf{83.2\%}      & 81.6\%             & 81.2\%                \\[7pt]
\begin{tabular}[c]{@{}c@{}}\textbf{MBPP}\\[-2pt] \scriptsize{ (pass@1)}\end{tabular}                  & 49.8\%               & 52.2\%             & \textbf{60.7\%}       \\[7pt]
\begin{tabular}[c]{@{}c@{}}\textbf{GSM-8K}\\[-2pt] \scriptsize{ (5-shot)}\end{tabular}                & 53.6\%               & 57.1\%             & \textbf{58.4\%}       \\[7pt]
\begin{tabular}[c]{@{}c@{}}\textbf{MT Bench}\\[-2pt] \scriptsize{ (for Instruct Models)}\end{tabular} & 6.86                 & \textbf{8.32}      & 8.30                  \\ \bottomrule
\end{tabular}
\vspace{4pt}
\caption{\small \textbf{Comparison of \mixtral with \llama 2 70B and GPT-3.5.} \mixtral outperforms or matches \llama 2 70B and GPT-3.5 performance on most metrics.}
\label{tab:vs_gpt35}
\vspace{-20pt}
\end{table}

\subsection{Multilingual benchmarks}
\vspace{-4pt}
Compared to \mistralSB, we significantly upsample the proportion of multilingual data during pretraining. The extra capacity allows \mixtral to perform well on multilingual benchmarks while maintaining a high accuracy in English.
In particular, \mixtral significantly outperforms \llama 2 70B in French, German, Spanish, and Italian, as shown in Table~\ref{tab:multilingual}.

\setlength{\tabcolsep}{2pt}

\begin{table}[h]
\centering
\scriptsize{
\begin{tabular}{@{}lcccccccccccccccccccc@{}}
\toprule
                      & \multirow{2}{*}{\textbf{\begin{tabular}[c]{@{}c@{}}Active \\ Params\end{tabular}}} & \textbf{} & \textbf{}       & \textbf{French} & \textbf{}       & \textbf{} & \textbf{} & \textbf{}       & \textbf{German} & \textbf{}       & \textbf{} & \textbf{} & \textbf{}       & \textbf{Spanish} & \textbf{}       & \textbf{} & \textbf{} & \textbf{}       & \textbf{Italian} & \textbf{}       \\
\textbf{Model}        &                                                                                    &           & Arc-c           & HellaS          & MMLU            &           &           & Arc-c           & HellaS          & MMLU            &           &           & Arc-c           & HellaS           & MMLU            &           &           & Arc-c           & HellaS           & MMLU            \\ \midrule
\textbf{LLaMA 1 33B}  & 33B                                                                                &           & 39.3\%          & 68.1\%          & 49.9\%          &           &           & 41.1\%          & 63.3\%          & 48.7\%          &           &           & 45.7\%          & 69.8\%           & 52.3\%          &           &           & 42.9\%          & 65.4\%           & 49.0\%          \\
\textbf{LLaMA 2 70B}  & 70B                                                                                &           & 49.9\%          & 72.5\%          & 64.3\%          &           &           & 47.3\%          & 68.7\%          & 64.2\%          &           &           & 50.5\%          & 74.5\%           & 66.0\%          &           &           & 49.4\%          & 70.9\%           & 65.1\%          \\
\textbf{\mixtralEXSB} & 13B                                                                                &           & \textbf{58.2\%} & \textbf{77.4\%} & \textbf{70.9\%} & \textbf{} & \textbf{} & \textbf{54.3\%} & \textbf{73.0\%} & \textbf{71.5\%} & \textbf{} & \textbf{} & \textbf{55.4\%} & \textbf{77.6\%}  & \textbf{72.5\%} & \textbf{} & \textbf{} & \textbf{52.8\%} & \textbf{75.1\%}  & \textbf{70.9\%} \\ \bottomrule
\end{tabular}}
\vspace{2pt}
\caption{
\small \textbf{Comparison of \mixtral with \llama on Multilingual Benchmarks.} On ARC Challenge, Hellaswag, and MMLU, \mixtral outperforms \llama 2 70B on 4 languages: French, German, Spanish, and Italian.
}
\vspace{-16pt}
\label{tab:multilingual}
\end{table}

\subsection{Long range performance}  
\begin{figure*}[h]
\vspace{-5pt}
\centering
\includegraphics[width=0.38\linewidth,height=\textheight,keepaspectratio,valign=t]{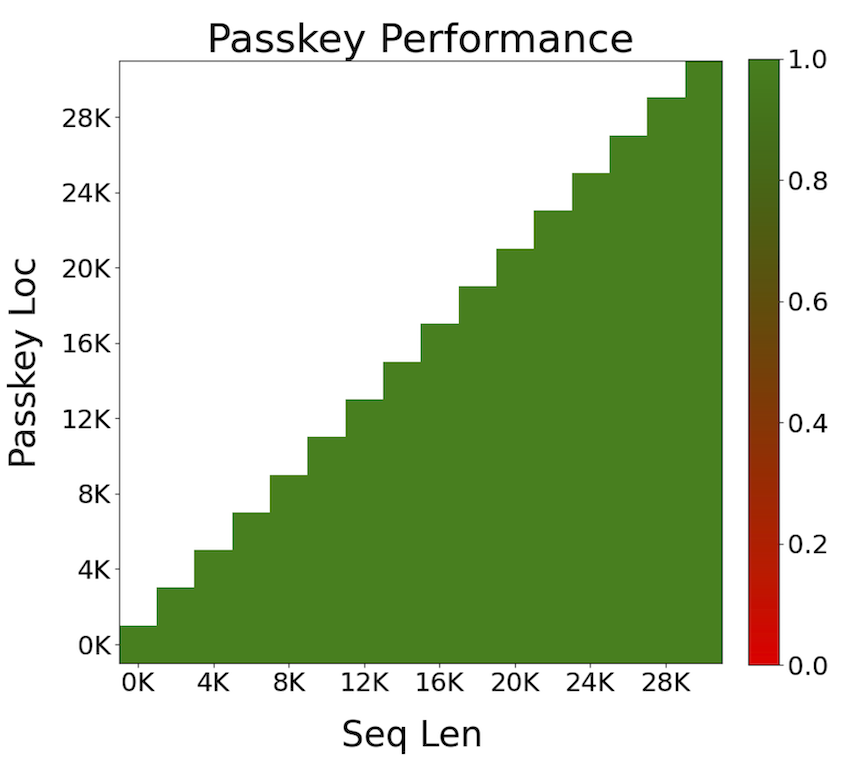}\hfill
\includegraphics[width=0.55\linewidth,height=\textheight,keepaspectratio, valign=t]{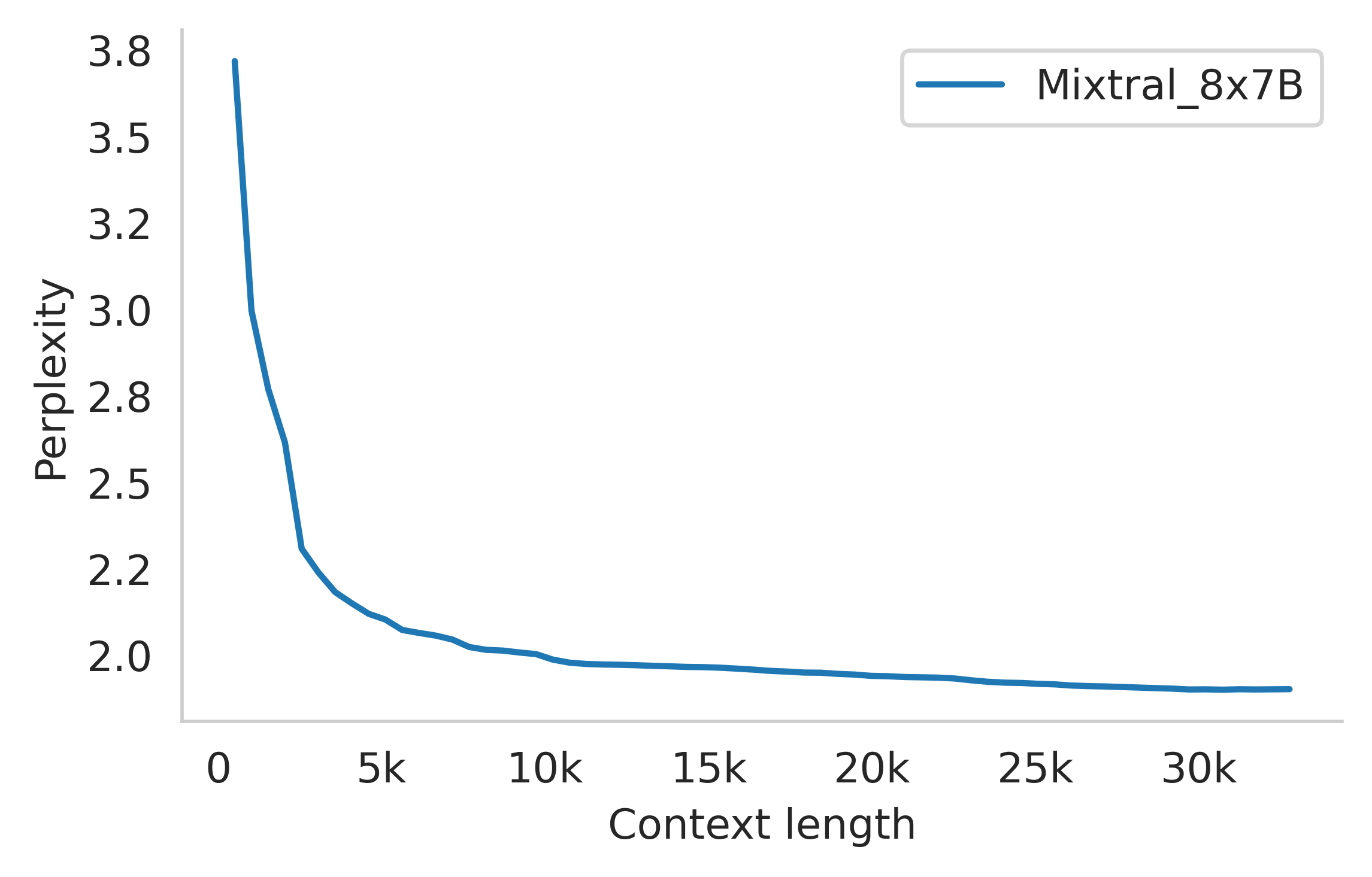}\hfill
\vspace{-8pt}
\caption{\small \textbf{Long range performance of \mixtral.} (Left) \mixtral has 100\% retrieval accuracy of the Passkey task regardless of the location of the passkey and length of the input sequence. (Right) The perplexity of \mixtral on the proof-pile dataset decreases monotonically as the context length increases.}
\label{fig:long_range}
\end{figure*}

\vspace{-4pt}
To assess the capabilities of \mixtral to tackle long context, we evaluate it on the passkey retrieval task introduced in~\cite{mohtashami2023landmark}, a synthetic task designed to measure the ability of the model to retrieve a passkey inserted randomly in a long prompt.
Results in Figure~\ref{fig:long_range} (Left) show that \mixtral achieves a 100\% retrieval accuracy regardless of the context length or the position of passkey in the sequence.
Figure~\ref{fig:long_range} (Right) shows that the perplexity of \mixtral on a subset of the proof-pile dataset~\cite{azerbayev2023llemma} decreases monotonically as the size of the context increases.

\pagebreak
\setlength{\tabcolsep}{10pt}
\begin{wrapfigure}{r}{0.54\textwidth}
\vspace{-10pt}
\small
\centering
    \begin{tabular}{@{}lcc@{}}
    \toprule
& \textbf{\llama 2 70B} & \textbf{\mixtralEXSB} \\ \midrule
    BBQ accuracy & 51.5\%              & 56.0\%  \\ 
    \midrule
\multicolumn{3}{l}{\hspace{-0.35cm}BOLD sentiment score (avg $\pm$ std) 
}
\\[3pt]  
    \footnotesize{gender}              & \footnotesize{0.293 $\pm$ 0.073}       & \footnotesize{0.323 $\pm$0.045}        \\[1pt]
    \footnotesize{profession}          & \footnotesize{0.218 $\pm$ 0.073}       & \footnotesize{0.243 $\pm$ 0.087}         \\[1pt]
    \footnotesize{religious\_ideology} & \footnotesize{0.188 $\pm$ 0.133}       & \footnotesize{0.144 $\pm$ 0.089}         \\[1pt]
    \footnotesize{political\_ideology} & \footnotesize{0.149 $\pm$ 0.140}       & \footnotesize{0.186 $\pm$ 0.146}         \\[1pt]
    \footnotesize{race}                & \footnotesize{0.232 $\pm$ 0.049}       & \footnotesize{0.232 $\pm$ 0.052}         \\ \bottomrule
    \end{tabular}
    \vspace{2pt}
    \caption{\small \textbf{Bias Benchmarks.} Compared \llama 2 70B, \mixtral presents less bias (higher accuracy on BBQ, lower std on BOLD) and displays more positive sentiment (higher avg on BOLD).}
    \label{tab:bias}
    \vspace{-10pt}
\end{wrapfigure}
\subsection{Bias Benchmarks} 

To identify possible flaws to be corrected by fine-tuning / preference modeling, we measure the base model performance on Bias Benchmark for QA (BBQ)~\cite{parrish2021bbq} and Bias in Open-Ended Language Generation Dataset (BOLD)~\cite{dhamala2021bold}.
BBQ is a dataset of hand-written question sets that target attested social biases against nine different socially-relevant categories: age, disability status, gender identity, nationality, physical appearance, race/ethnicity, religion, socio-economic status, sexual orientation. BOLD is a large-scale dataset that consists of 23,679 English text generation prompts for bias benchmarking across five domains.

\looseness=-1 We benchmark \llama 2 and \mixtral on BBQ and BOLD with our evaluation framework and report the results in Table~\ref{tab:bias}. Compared to \llama 2, \mixtral presents less bias on the BBQ benchmark (56.0\% vs 51.5\%). For each group in BOLD, a higher average sentiment score means more positive sentiments and a lower standard deviation indicates less bias within the group. Overall, \mixtral displays more positive sentiments than \llama 2, with similar variances within each group.

\section{Instruction Fine-tuning}
\begin{figure*}[h]
\centering
\includegraphics[width=0.9\linewidth,height=\textheight,keepaspectratio]{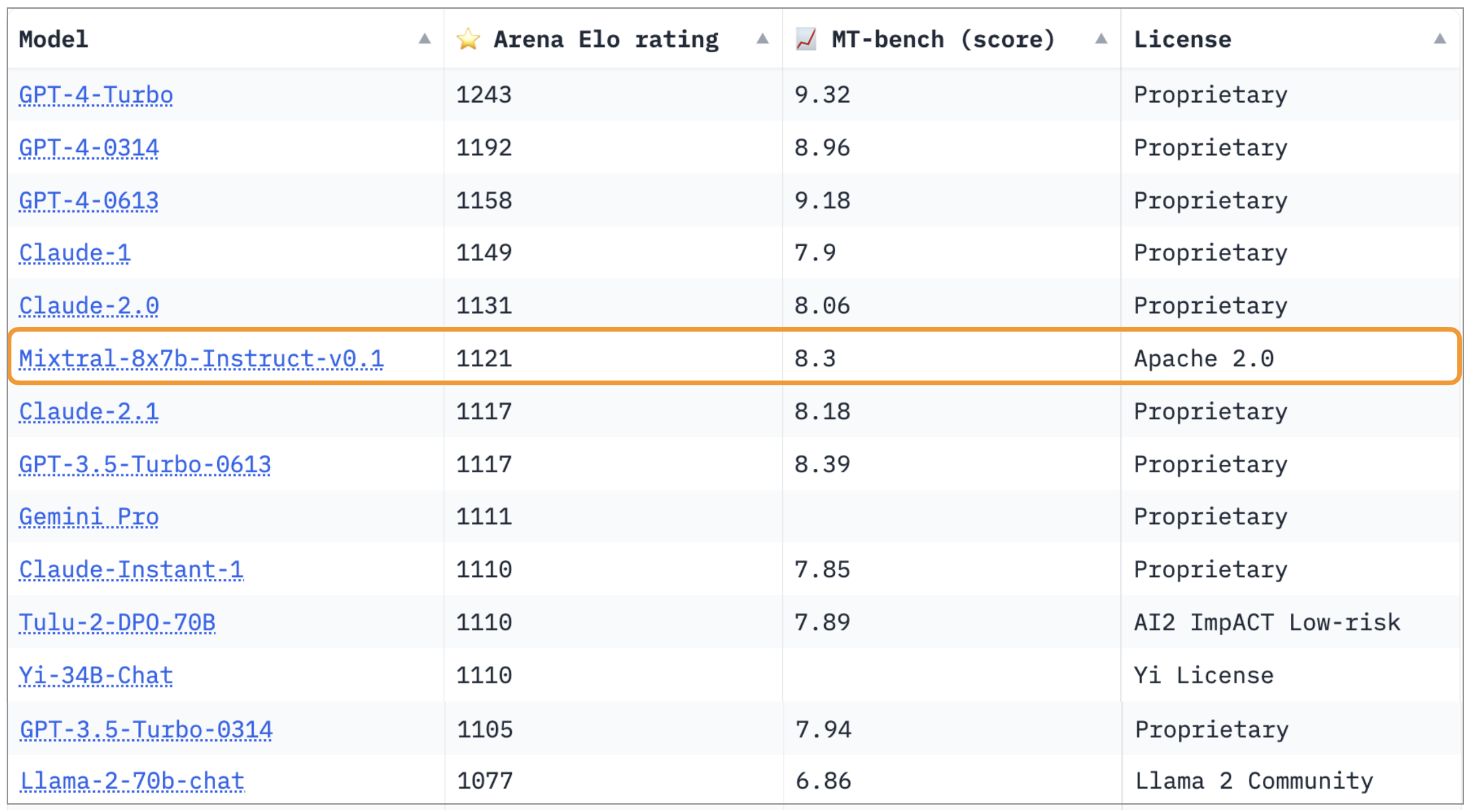}
\caption{
\small \textbf{LMSys Leaderboard.} (Screenshot from Dec 22, 2023) \mixtralEXSB Instruct v0.1 achieves an Arena Elo rating of 1121 outperforming Claude-2.1 (1117), all versions of GPT-3.5-Turbo (1117 best), Gemini Pro (1111), and Llama-2-70b-chat (1077). \mixtral is currently the best open-weights model by a large margin.
}
\label{fig:lmsys}
\end{figure*}

\looseness=-1 We train \mixtralchat using supervised fine-tuning (SFT) on an instruction dataset followed by Direct Preference Optimization (DPO)~\cite{rafailov2023direct} on a paired feedback dataset. \mixtralchat reaches a score of 8.30 on MT-Bench~\cite{zheng2023judging} (see Table~\ref{tab:results}), making it the best open-weights model as of December 2023.
Independent human evaluation conducted by LMSys is reported in Figure~\ref{fig:lmsys}\footnote{\url{https://huggingface.co/spaces/lmsys/chatbot-arena-leaderboard}} and shows that \mixtralchat outperforms GPT-3.5-Turbo, Gemini Pro, Claude-2.1, and \llama~2~70B chat.

\section{Routing analysis}
\vspace{-5pt}

In this section, we perform a small analysis on the expert selection by the router.
In particular, we are interested to see if during training some experts specialized to some specific domains (e.g. mathematics, biology, philosophy, etc.).

\begin{figure}[h]
\centering
\includegraphics[width=1\linewidth,keepaspectratio]{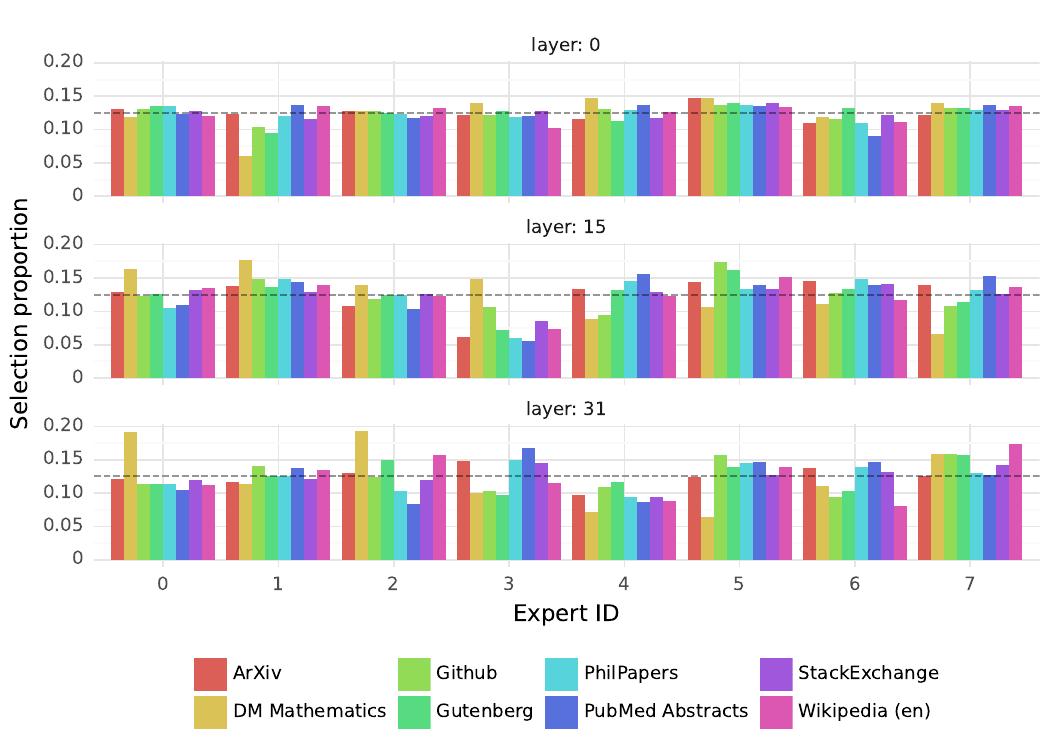}
\caption{
\looseness=-1 \small \textbf{Proportion of tokens assigned to each expert on different domains from The Pile dataset for layers 0, 15, and 31.}
The gray dashed vertical line marks $1/8$, i.e. the proportion expected with uniform sampling.
Here, we consider experts that are either selected as a first or second choice by the router.
A breakdown of the proportion of assignments done in each case cane be seen in Figure~\ref{fig:smoeroutingassignmentfull} in the Appendix.
}
\label{fig:smoeroutingassignment}
\end{figure}

To investigate this, we measure the distribution of selected experts on different subsets of The Pile validation dataset~\cite{gao2020pile}.
Results are presented in Figure~\ref{fig:smoeroutingassignment}, for layers 0, 15, and 31 (layers 0 and 31 respectively being the first and the last layers of the model).
Surprisingly, we do not observe obvious patterns in the assignment of experts based on the topic.
For instance, at all layers, the distribution of expert assignment is very similar for ArXiv papers (written in Latex), for biology (PubMed Abstracts), and for Philosophy (PhilPapers) documents.

Only for DM Mathematics we note a marginally different distribution of experts.
This divergence is likely a consequence of the dataset's synthetic nature and its limited coverage of the natural language spectrum, and is particularly noticeable at the first and last layers, where the hidden states are very correlated to the input and output embeddings respectively.

\begin{figure}[b]
\centering
\includegraphics[width=\linewidth,keepaspectratio]{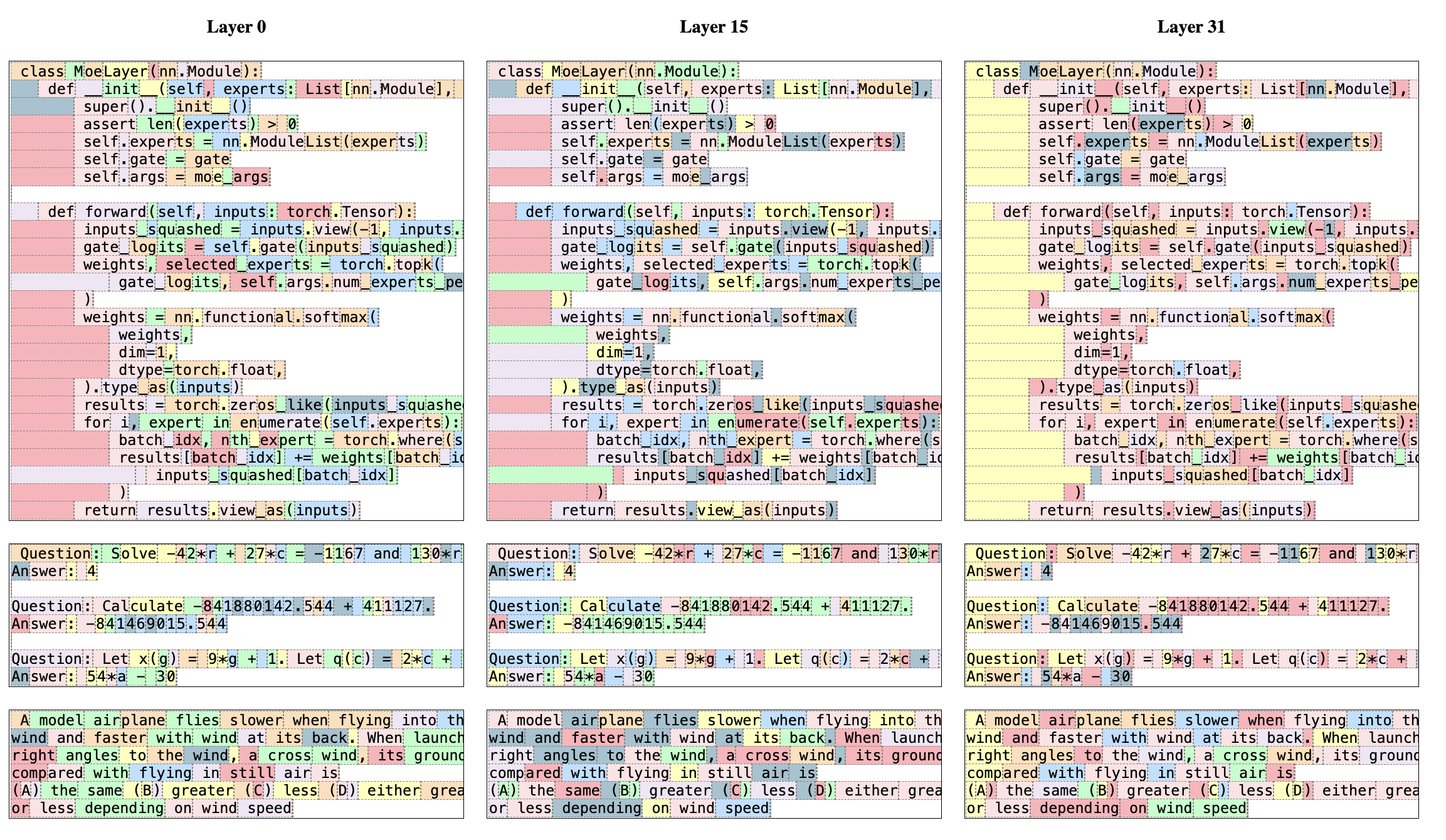}
\caption{\small \textbf{Text samples where each token is colored with the first expert choice.}
The selection of experts appears to be more aligned with the syntax rather than the domain, especially at the initial and final layers.
}
\label{fig:smoecoloredtext}
\end{figure}

This suggests that the router does exhibit some structured syntactic behavior.
Figure~\ref{fig:smoecoloredtext} shows examples of text from different domains (Python code, mathematics, and English), where each token is highlighted with a background color corresponding to its selected expert.
The figure shows that words such as `self' in Python and `Question' in English often get routed through the same expert even though they involve multiple tokens.
Similarly, in code, the indentation tokens are always assigned to the same experts, particularly at the first and last layers where the hidden states are more correlated to the input and output of the model.

\looseness=-1 We also note from Figure~\ref{fig:smoecoloredtext} that consecutive tokens are often assigned the same experts. In fact, we observe some degree of positional locality in The Pile datasets. Table~\ref{tab:smoerepeat} shows the proportion of consecutive tokens that get the same expert assignments per domain and layer. The proportion of repeated 

\looseness=-1 consecutive assignments is significantly higher than random for higher layers. This has implications in how one might optimize the model for fast training and inference. For example, cases with high locality are more likely to cause over-subscription of certain experts when doing Expert Parallelism. Conversely, this locality can be leveraged for caching, as is done in \cite{eliseev2023fast}.
A more complete view of these same expert frequency is provided for all layers and across datasets in Figure~\ref{fig:smoerepeated} in the Appendix.

\begin{table}
\small
\centering

\begin{tabular}{l|ccc|ccc}
\toprule
& \multicolumn{3}{c}{First choice} & \multicolumn{3}{c}{First or second choice} \\
 & Layer 0 & Layer 15 & Layer 31 & Layer 0 & Layer 15 & Layer 31 \\
\midrule
ArXiv & 14.0\% & 27.9\% & 22.7\% & 46.5\% & 62.3\% & 52.9\% \\
DM Mathematics & 14.1\% & 28.4\% & 19.7\% & 44.9\% & 67.0\% & 44.5\% \\
Github & 14.9\% & 28.1\% & 19.7\% & 49.9\% & 66.9\% & 49.2\% \\
Gutenberg & 13.9\% & 26.1\% & 26.3\% & 49.5\% & 63.1\% & 52.2\% \\
PhilPapers & 13.6\% & 25.3\% & 22.1\% & 46.9\% & 61.9\% & 51.3\% \\
PubMed Abstracts & 14.2\% & 24.6\% & 22.0\% & 48.6\% & 61.6\% & 51.8\% \\
StackExchange & 13.6\% & 27.2\% & 23.6\% & 48.2\% & 64.6\% & 53.6\% \\
Wikipedia (en) & 14.4\% & 23.6\% & 25.3\% & 49.8\% & 62.1\% & 51.8\% \\
\bottomrule
\end{tabular}

\vspace{10pt}
\caption{
\small
\textbf{Percentage of expert assignment repetitions.}
\looseness=-1 We evaluate the proportion of times the same expert is assigned to a token $i$ and its following token $i+1$. We report whether the first chosen expert is the same, or whether the same expert is observed as first or second choice in consecutive tokens.
For reference, the expected proportion of repetitions in the case of random assignments is $\frac{1}{8}=12.5\%$ for ``First choice'' and $1-\frac{6}{8} \frac{5}{7} \approx 46\%$ for ``First and second choice''.
Repetitions at the first layer are close to random, but are significantly higher at layers 15 and 31.
The high number of repetitions shows that expert choice exhibits high temporal locality at these layers. 
}
\vspace{-10pt}
\label{tab:smoerepeat}
\end{table}

\section{Conclusion}
\vspace{-5pt}

\looseness=-1 In this paper, we introduced \mixtralEXSB, the first mixture-of-experts network to reach a state-of-the-art performance among open-source models.
\mixtralEXSB Instruct outperforms Claude-2.1, Gemini~Pro, and GPT-3.5~Turbo on human evaluation benchmarks.
Because it only uses two experts at each time step, \mixtral only uses 13B active parameters per token while outperforming the previous best model using 70B parameters per token (\llama~2~70B).
We are making our trained and fine-tuned models publicly available under the Apache 2.0 license.
By sharing our models, we aim to facilitate the development of new techniques and applications that can benefit a wide range of industries and domains.

\pagebreak

\section*{Acknowledgements}

We thank the CoreWeave and Scaleway teams for technical support as we trained our models.
We are grateful to NVIDIA for supporting us in integrating TensorRT-LLM and Triton and working alongside us to make a sparse mixture of experts compatible with TensorRT-LLM.

\bibliographystyle{plain}
\bibliography{ref}

\appendix

\begin{figure*}
\centering
\includegraphics[width=0.9\linewidth]{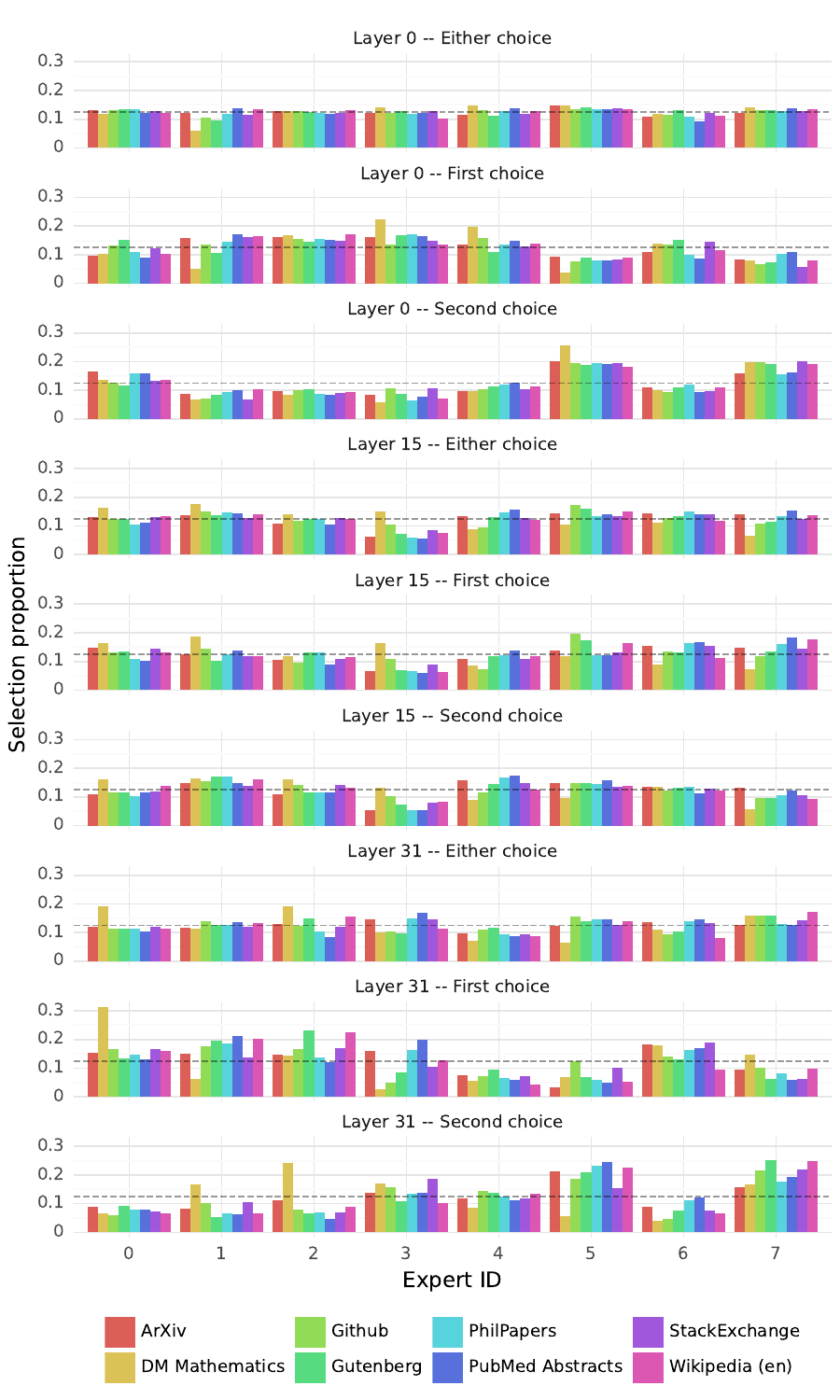}
\caption{
\looseness=-1 \small \textbf{Proportion of tokens assigned to each expert on different subsets from The Pile dataset, separated by whether the expert was selected as first or second choice, or either.} The ``Either choice'' case is equivalent to Figure~\ref{fig:smoeroutingassignment}.
The gray dashed vertical line marks $\frac{1}{8}$, i.e. the proportion expected with uniform sampling.
}
\label{fig:smoeroutingassignmentfull}
\end{figure*}

\begin{figure*}
\centering
\includegraphics[width=0.99\linewidth]{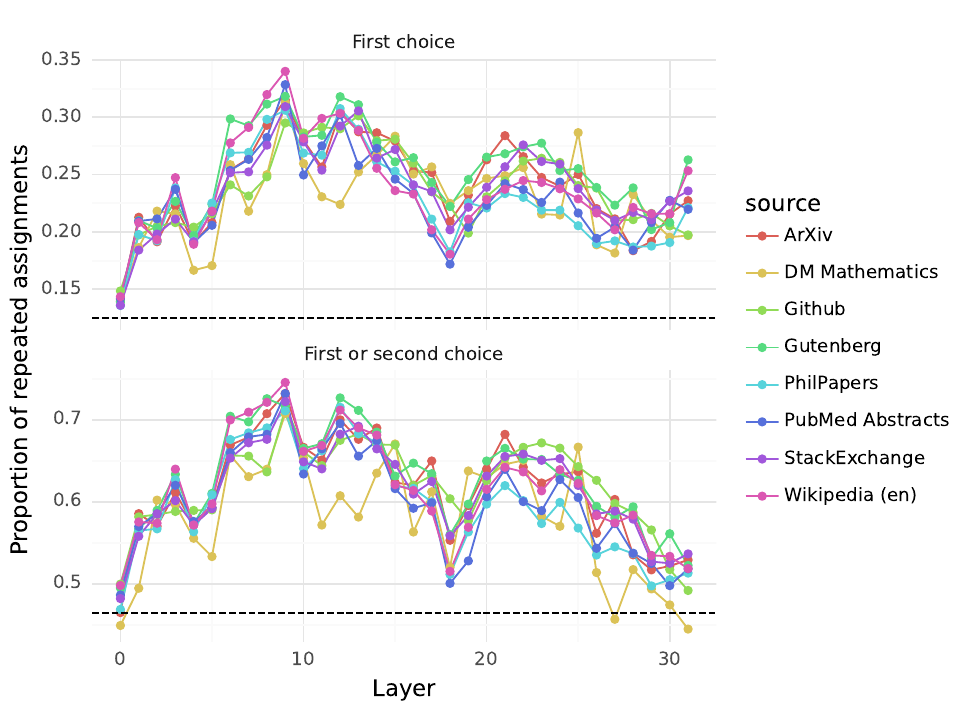}
\caption{
\looseness=-1 \small \textbf{Repeated consecutive assignments per MoE layer.} Repeated assignments occur a lot more often than they would with uniform assignments (materialized by the dashed lines). Patterns are similar across datasets with less repetitions for DM Mathematics.
}
\label{fig:smoerepeated}
\end{figure*}
\end{document}